\documentclass{article}

\usepackage{arxiv}

\usepackage[utf8]{inputenc} 
\usepackage[T1]{fontenc}    
\usepackage{hyperref}       
\usepackage{url}            
\usepackage{booktabs}       
\usepackage{amsfonts}       
\usepackage{nicefrac}       
\usepackage{microtype}      
\usepackage{lipsum}		
\usepackage{graphicx}
\usepackage{doi}
\usepackage{placeins} 
\usepackage{amsmath}
\usepackage[ruled,vlined]{algorithm2e}
\usepackage{bbm,amsmath}
\usepackage{comment}
\usepackage{enumitem}
\usepackage{siunitx}
\usepackage{bbold}
\usepackage{mathtools}
\usepackage[dvipsnames]{xcolor}

\title{On Evolvability and Behavior Landscapes in Neuroevolutionary Divergent Search}

\author{ \href{https://orcid.org/0000-0003-0486-233X}{\includegraphics[scale=0.06]{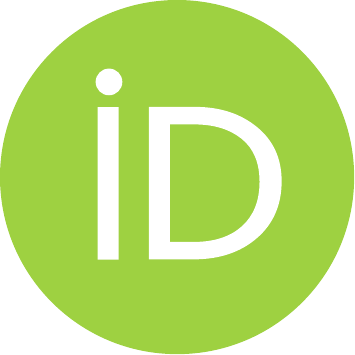}\hspace{1mm}Bruno Gašperov} \\
        University of Zagreb, Faculty of \\
        Electrical Engineering and Computing \\
        Zagreb, Croatia \\
	\texttt{bruno.gasperov@fer.hr} \\
	\And
	\href{https://orcid.org/0000-0001-8732-4769}{\includegraphics[scale=0.06]{orcid.pdf}\hspace{1mm}Marko Đurasević} \\
        University of Zagreb, Faculty of \\
        Electrical Engineering and Computing \\
        Zagreb, Croatia \\
	\texttt{marko.durasevic@fer.hr} \\
}

\date{}

\renewcommand{\shorttitle}{On Evolvability and Behavior Landscapes in Neuroevolutionary Divergent Search}

\hypersetup{
pdftitle={A Search for Nonlinear Balanced Boolean Functions by
Leveraging Phenotypic Properties},
pdfsubject={q-bio.NC, q-bio.QM},
pdfauthor={David S.~Hippocampus, Elias D.~Striatum},
pdfkeywords={First keyword, Second keyword, More},
}

\begin{document}
\maketitle

\begin{abstract}
Evolvability refers to the ability of an individual genotype (solution) to produce offspring with mutually diverse phenotypes. Recent research has demonstrated that divergent search methods, particularly novelty search, promote evolvability by implicitly creating selective pressure for it. The main objective of this paper is to provide a novel perspective on the relationship between neuroevolutionary divergent search and evolvability. In order to achieve this, several types of walks from the literature on fitness landscape analysis are first adapted to this context. Subsequently, the interplay between neuroevolutionary divergent search and evolvability under varying amounts of evolutionary pressure and under different diversity metrics is investigated. To this end, experiments are performed on Fetch Pick and Place, a robotic arm task. Moreover, the performed study in particular sheds light on the structure of the genotype-phenotype mapping (the behavior landscape). Finally, a novel definition of evolvability that takes into account the evolvability of offspring and is appropriate for use with discretized behavior spaces is proposed, together with a Markov-chain-based estimation method for it.
\end{abstract}


\section{Introduction}

Recent years have seen a strong surge of interest in a number of evolutionary computation and population-based search approaches rooted in the ideas of divergent search and open-endedness \cite{stanley2019open}. When combined with neuroevolution, these approaches offer a promising alternative \cite{such2017deep, salimans2017evolution} to more canonical gradient-based algorithms for training deep reinforcement learning agents, especially for tasks that include sparse or deceptive rewards, or require generation of a large number of high-performing yet behaviorally diverse solutions. Within this realm, the novelty search \cite{lehman2011novelty} and quality diversity \cite{pugh2016quality} families of approaches have gained considerable attention in the research community. Novelty search is a class of divergent search algorithms based on abandoning objectives altogether and solely pursuing behavioral novelty instead. Novelty is typically evaluated by comparing the behavior of an individual to the behavior of the current population and a record of previously encountered individuals, known as the archive. On the other hand, quality-diversity approaches (as the name itself implies) consider both the objective quality of solutions and their diversity, aiming to discover a diverse set of high-performing solutions that occupy different niches, each of which represents a different type of behavior. As these algorithms "illuminate" the fitness potential of feature space areas, they are sometimes referred to as illumination algorithms \cite{mouret2015illuminating}. The underlying thread of all divergent search algorithms is their emphasis on enabling the necessary conditions for open-ended processes to occur, which requires a high level of what is known as "evolvability".

The concept of evolvability has been frequently studied, both in the area of natural evolution \cite{dawkins2019evolution} and evolutionary computing \cite{wilder2015reconciling, gajewski2019evolvability}. Despite the abundance of various, even somewhat conflicting definitions \cite{pigliucci2008evolvability}, we see evolvability as the capacity of an individual (genotype) to produce offspring with diverse phenotypes. Following this, evolvability can be considered an indicator of the individual's ability to exhaustively explore the solution space through its offspring. While the importance of evolvability for divergent search can hardly be overstated, given that it increases the potential for future evolution \cite{mengistu2016evolvability}, this area of research remains relatively unexplored, especially in combination with neuroevolution. It has been shown that novelty search in itself promotes evolvability \cite{lehman2016critical} (measured via reachability and sampling uniformity of the offspring) on both individual and population level \cite{doncieux2020novelty}, with some caveats\footnote{A distinction should be made between novelty search for neuroevolution of weights and neuroevolution of topologies.} \cite{lehman2011improving, cuccu2011novelty,shorten2014evolvable}. Yet novelty search boils down to greedily choosing the most diverse individuals within each generation without taking into account their other characteristics, including evolvability itself, which might lead to less diversity in the long run. Similar considerations have led researchers \cite{mengistu2016evolvability, gajewski2019evolvability} to propose algorithms that instead search directly for evolvability. The successes of these approaches in generalizing well and producing higher evolvability even in unseen testing environments, as well as in yielding solutions that can be quickly adapted for different tasks, are demonstrated. A somewhat different path is taken by Katona \emph{et al.} \cite{katona2021quality} who propose the Quality Evolvability Evolutionary Strategy. In their work, a population of diverse and well-performing solutions is obtained indirectly - by finding a single individual with such a distribution of offspring. In \cite{doncieux2020novelty}, the authors list several properties, characteristics, and mechanisms which were demonstrated to be conducive to improving evolvability, including extinction events \cite{lehman2015extinction} and reduction of the cost of connections between network nodes \cite{clune2013evolutionary}. Ferigo \emph{et al.} \cite{ferigo2022entanglement} first define evolvability as the mean difference in fitness between the parents and their offspring, and then use a grid-like structure for studying the interplay between evolvability and fitness. Their main findings suggest that evolvability is not an intrinsic property of the fitness landscape, but that it rather heavily depends on the used evolutionary algorithm. Finally, we mention the newer work of Doncieux \emph{et al.} \cite{doncieux2019novelty} who treat evolvability as reachability in the task space and provide a theoretical perspective on the underlying problem.

Our research lies at the intersection of (behavior) landscape analysis, neural network weight neuroevolution, and evolvability research, exploring the strong links between the three topics. Following definitions from \cite{doncieux2020novelty}, we are primarily concerned with individual evolvability, measured via a form of reachability, as opposed to population evolvability \cite{wilder2015reconciling, lehman2016critical}. Furthermore, unlike the authors in \cite{ferigo2022entanglement}, we study behavior for its own sake and do not rely on any particular fitness function. We particularly explore the effect of the use of different diversity metrics and levels of evolutionary pressure on evolvability in the context of neuroevolutionary divergent search. 

The main contributions of this study are the following. First, we introduce a number of metrics related to evolvability and behavior landscape analysis. Then we adapt the concept of walks from the existing literature on fitness landscape analysis to the context of behavior landscape analysis. We also propose "dissimila", points with low evolvability that play a role in novelty search that we suspect is similar to that of local optima in standard fitness-driven search. Second, we propose a diversity metric based on the use of Gaussian kernel density estimates. Third, we study the sensitivity of individual evolvability to the level of evolutionary pressure. By varying the level of evolutionary pressure present in the environment, we aim to generalize existing research on the connections between novelty search and evolvability. Positive links between evolvability and the level of evolutionary pressure present in the environment would provide additional evidence on the significance of novelty search for promoting evolvability. Fourth, we investigate the effect of different diversity metrics on evolvability, with a focus on the aforementioned kernel density estimation-based metric. A suitable diversity metric is a key heuristic in a divergent search algorithm as it plays a vital role in assessing the diversification of the solutions produced by the algorithm. Fifth and finally, we put forth a novel definition of evolvability that also considers the evolvability of offspring and is suitable for use with discretized behavior spaces, together with an associated Markov-chain-based estimation method.

The paper is organized as follows. Section 2 introduces definitions and metrics which represent the prerequisites for further work. In particular, concepts such as evolvability, sensitivity, and walks are expounded on in great detail, and a number of diversity metrics are listed. Section 3 presents the experimental setup, including the task description, operators, neural network architecture, and hyperparameters. In Section 4, the experimental results are presented. Section 5 proposes a novel evolvability definition together with an associated estimation method. Finally, Section 6 concludes the paper by summarizing the findings and offering suggestions for future research in the field.

\section{Definitions and metrics}
\label{chapter:2}

\subsection{Behavior function and descriptors}

A behavior function $\phi : \Theta \mapsto \mathcal{B}$ is a mapping from genotypes (in neuroevolution represented by parameters of a neural network) to phenotypes (behavior vectors):
\begin{equation}
\phi(\boldsymbol{\theta}) = \mathbf{b},
\end{equation}
where $\Theta$ is the set of genotypes, $\mathcal{B}$ is the behavior space (BS), i.e., the set of all behavior vectors (BVs), $\boldsymbol{\theta}$ is a genotype, and $\mathbf{b}$ its phenotype. Typically, $\mathcal{B} \subseteq \mathbb{R}^n$ with $n \geq 2$. The components of a BV are referred to as behavior descriptors. For example, a common choice of behavior descriptors in evolutionary robotics is given by the final position of the robot's body along two dimensions, i.e., $\mathbf{b} = (x_T, y_T)$ where $T$ is the terminal time. In some approaches (e.g., the MAP-Elites algorithm \cite{mouret2015illuminating}), the BS is split into niches - regions or subspaces containing solutions with similar behavior, which can be obtained by simply discretizing the BS into a grid. As an alternative, centroidal Voronoi tessellation \cite{vassiliades2017using}, which ensures scalability to high-dimensional BSs, might be used. The use of niches gives rise to an alternative categorical representation of BVs in the form of one-hot vectors $\mathbf{b} = (\ldots, \mathbb{1}_{\text{behavior in the niche i}}, \ldots )$, with the $i$-th entry set to $1$ if the BV belongs to the $i$-th niche, and $0$ otherwise. However, it is important to notice that this encoding discards the original relationships between the niches and eliminates any naturally occurring distance metrics between them.

\subsection{Behavior landscape}

A behavior landscape is an abstraction that represents the genotype-behavior mapping, i.e., the function $\phi$. We define it analogously to the fitness landscape, which deals with genotype-fitness mapping and has been thoroughly studied in various contexts, and the behavior-fitness mapping, considered widely in quality-diversity approaches. This triad of mappings forms a bridge between classical evolutionary algorithms and new illumination-based approaches. Behavior landscapes are intricate and complex structures influenced by a number of factors, such as the choice of the behavior function, the topology of the (neural network) controller\footnote{A controller simply determines the mapping from the set of observations (states) to the set of actions, which can be represented by a reinforcement learning agent.}, the interaction of the resulting controls with the environment, and the environment's own internal mechanics. They are typically highly rugged, showing significant epistasis \cite{davidor1991epistasis}, and more difficult to visualize than fitness landscapes \cite{pitzer2012comprehensive}, given the possible high dimensionality of the BS. A fitness landscape is usually considered difficult if it has a large number of peaks and valleys (local optima) or if the paths to global optima are not easy to traverse. In behavior landscape analysis, different criteria are required, and ruggedness is not necessarily a hindrance, as it is closely related to diversity. A difficult behavior landscape could be seen as one in which a divergent search algorithm struggles to explore the BS easily due to the presence of inaccessible regions. Therefore, different topographical aspects (compared to classical fitness landscape analysis), such as the distribution of different niches in the behavior landscape, should be scrutinized. On a related note, we remark that some studies suggest that behaviorally diverse solutions lie concentrated in a small subset of the genotype space called the hypervolume\footnote{In \cite{vassiliades2018discovering}, the emphasis is put on elite hypervolumes in the sense of quality-diversity.} \cite{vassiliades2018discovering, rakicevic2021policy, gaier2020discovering}. 

It has also been noted \cite{doncieux2020novelty} that, in the context of novelty search, the fitness function is dynamic and contingent upon the set of individuals comprising the current population and archive. In contrast, the behavior landscape is static\footnote{In stochastic environments, where behaviors are noisy, behavior descriptors (which are now random variables) can be replaced by their expected values, thereby preserving the static nature of the behavior function.}, rendering it less recalcitrant to analysis. Yet the research area remains mainly understudied with some notable exceptions \cite{salehi2022geodesics}, while even the broader domain of neuroevolutionary fitness landscapes has received very limited scholarly attention \cite{rodrigues2020study}.

\subsection{Local sensitivity}

A behavior function $\phi$ is said to be locally sensitive at $\boldsymbol{\theta}$ if small steps around it can lead to large changes in the ensuing behavior. More formally, the \textbf{local sensitivity} of a behavior function $\phi$ at $\boldsymbol{\theta} \in \Theta$ can be defined as:
\begin{equation} \label{eq:1}
    LS_{\phi}(\boldsymbol{\theta}) = \underset{\boldsymbol{\theta'} \in \mathcal{N}(\boldsymbol{\theta}) }{\mathrm{max}} \ |\phi(\boldsymbol{\theta})-\phi(\boldsymbol{\theta'})|,
\end{equation}
where $\mathcal{N}(\boldsymbol{\theta})$ is the set of all neighbours of $\boldsymbol{\theta}$, associated with a certain mutation operator $\mathcal{M}$. $\boldsymbol{\theta'}$ is a neighbor of $\boldsymbol{\theta}$ if and only if $\boldsymbol{\theta'}$ can be reached from $\boldsymbol{\theta}$ through a single mutation. This relation is not necessarily symmetric\footnote{E.g., consider removal of a neural network weight as a mutation operator.}. Large $LS_{\phi}(\boldsymbol{\theta})$ values point to the existence of neighbors that are significantly behaviorally different from $\boldsymbol{\theta}$ but say nothing about their frequency. On the contrary, small $LS_{\phi}(\boldsymbol{\theta})$ values indicate high levels of behavioral neutrality \cite{galvan2021neuroevolution} locally around $\boldsymbol{\theta}$. Note that, when training neural networks via neuroevolution, mutations are often given as random samples from a multivariate (normal) distribution used for perturbing its weights, in which case $\mathcal{N}(\boldsymbol{\theta}) = \Theta$, rendering the notion of the set of neighbors meaningless. However, this can be easily addressed by truncating the underlying distribution or by otherwise enforcing genotypic proximity given by $|\boldsymbol{\theta}-\boldsymbol{\theta'}| < c$, for some positive constant $c$. As previously hinted at, Eq. \ref{eq:1} is sensitive to outliers due to the use of the \verb|max| operator. A more robust alternative can be provided. First, consider the sampling distribution over the set of neighbors $\mathcal{N}(\boldsymbol{\theta})$ induced by $\mathcal{M}$ and denote it by $\mathcal{F}$. Now simply define the \textbf{expected local sensitivity} as:
\begin{equation} \label{eq:kik}
    LS_{\phi}^{*}(\boldsymbol{\theta}) = \underset{\boldsymbol{\theta'} \sim \mathcal{F}(\boldsymbol{\theta'} \in \mathcal{N}(\boldsymbol{\theta})) }{\mathbb{E}}|\phi(\boldsymbol{\theta})-\phi(\boldsymbol{\theta'})|.
\end{equation}
Eq. \ref{eq:kik} measures the expected distance between $\boldsymbol{\theta}$ and its randomly selected neighbor. It can be estimated by sampling and taking the mean value:
\begin{equation} \label{eq:2.5}
    \overline{LS_{\phi}^{*}(\boldsymbol{\theta})} = \frac{1}{\mathbf{card}(\mathcal{S}(\boldsymbol{\theta}))} \sum_{\boldsymbol{\boldsymbol{\theta'}} \in \mathcal{S}(\boldsymbol{\theta})} |\phi(\boldsymbol{\theta})-\phi(\boldsymbol{\theta'})|,
\end{equation}
where $\mathcal{S}(\boldsymbol{\theta}) \subseteq \mathcal{N}(\boldsymbol{\theta})$ is a sample from the set of neighbours of $\boldsymbol{\theta}$ and $\mathbf{card}(\ldots)$ denotes cardinality.


\subsection{Global sensitivity} The \textbf{global sensitivity} of $\phi$ is defined as:
\begin{equation}
    GS_{\phi} = \underset{\boldsymbol{\theta}' \in \Theta , \ \boldsymbol{\theta''} \in \mathcal{N}(\boldsymbol{\theta}') }{\mathrm{max}} \ |\phi(\boldsymbol{\theta'})-\phi(\boldsymbol{\theta''})|.
\end{equation}
It is equal to the local sensitivity of the genotype $\boldsymbol{\theta}$ that has the largest local sensitivity. The \textbf{expected global sensitivity} is defined in analogy to Eq. \ref{eq:kik} but is not provided here for the sake of brevity.

\subsection{Evolvability} \textbf{Evolvability} \cite{doncieux2020novelty} can be conceptualized as a gauge of an individual's capacity to produce offspring with mutually diverse phenotypes (in our case BVs). Although there is no consensus on its definition, we use the following one:
\begin{equation} \label{eq:1.5}
    \eta_{\phi}(\boldsymbol{\theta}) = \underset{\boldsymbol{\theta'}, \ \boldsymbol{\theta''} \in \mathcal{N}(\boldsymbol{\theta}) }{\mathrm{max}} \ |\phi(\boldsymbol{\theta'})-\phi(\boldsymbol{\theta''})|,
\end{equation}
where $\eta_{\phi}(\boldsymbol{\theta})$ denotes the evolvability of solution $\boldsymbol{\theta}$. Hence, evolvability corresponds to the diameter of the cluster comprising the neighbors of $\boldsymbol{\theta}$. The same caveats that apply to Eq. \ref{eq:1} apply here as well. By comparing Eq. \ref{eq:1} and Eq. \ref{eq:1.5} it is clear that, due to the triangle inequality, $\eta_{\phi}(\boldsymbol{\theta}) \leq 2\ LS_{\phi} (\boldsymbol{\theta})$, and the two metrics are expected to be positively correlated. Again, similarly to Eq. \ref{eq:kik}, the \textbf{expected evolvability} is a robust alternative given by:
\begin{equation}
    \eta_{\phi}^{*}(\boldsymbol{\theta}) = \underset{\boldsymbol{\theta'}, \ \boldsymbol{\theta''} \sim \mathcal{F}(\boldsymbol{\theta'}, \ \boldsymbol{\theta''} \in \mathcal{N}(\theta))}{\mathbb{E}}|\phi(\boldsymbol{\theta}')-\phi(\boldsymbol{\theta}'')|
\end{equation}
and its estimator:
\begin{equation}
    \overline{\eta_{\phi}^{*}(\boldsymbol{\theta})} = \frac{1}{\mathbf{card}(\mathcal{S}(\boldsymbol{\theta}))\ \mathbf{card}(\mathcal{S}(\boldsymbol{\theta})-1)} \sum_{\substack{\boldsymbol{\theta'}, \  \boldsymbol{\theta''} \in \mathcal{S}(\boldsymbol{\theta})\\ \boldsymbol{\theta'} \neq \boldsymbol{\theta''}}} |\phi(\boldsymbol{\theta}')-\phi(\boldsymbol{\theta}'')|.
\end{equation}
Under a one-hot representation of BVs, evolvability can be estimated relative to the sample size, as: 
\begin{equation}
    \overline{\eta_{\phi}(\boldsymbol{\theta}; \mathbf{card}({\mathcal{S}(\boldsymbol{\theta})))}} = \frac{d}{n},
\end{equation}
where $n$ is the total number of niches, and $d$ the number of niches occupied by solutions from the sample $\mathcal{S}(\boldsymbol{\theta})$. Although our focus is on individual-level evolvability, we also present the following expression used to estimate the \textbf{expected population evolvability}:

\begin{equation}
    \overline{\rho_{\phi}^{*}(i)} = \frac{1}{n_{i+1} (n_{i+1}-1)} \sum_{\substack{\boldsymbol{\theta'}, \  \boldsymbol{\theta''} \in \mathcal{G}_{i+1}\\ \boldsymbol{\theta'} \neq \boldsymbol{\theta''}}} |\phi(\boldsymbol{\theta}')-\phi(\boldsymbol{\theta}'')|,
\end{equation}
where $\rho_{\phi}^{*}(i)$ denotes the expected population evolvability of the $i$-th generation, $\mathcal{G}_{i}$ the set of all members of the $i$-th evolutionary generation, and $n_{i}$ its cardinality. Finally, it is worth noting that evolvability is generally computationally expensive, and in some cases, even prohibitively so, since it requires evaluating the fitness function (which can itself be costly) on the whole $\mathcal{N}(\boldsymbol{\theta})$ or a sufficiently large sample of it.

\subsection{Local "optima"}

Consider the ratio of evolvability and local sensitivity:
\begin{equation}
    r(\boldsymbol{\theta}) = \frac{\eta_{\phi}(\boldsymbol{\theta})}{LS_{\phi}(\boldsymbol{\theta})+\epsilon},
\end{equation}
where protected division is employed (for some infinitesimal $\epsilon>0$). It can be shown that $0 \leq r(\boldsymbol{\theta}) < 2$. Also, consider its probabilistic variant:
\begin{equation}
    r^{*}(\boldsymbol{\theta}) = \frac{\eta_{\phi}^{*}(\boldsymbol{\theta})}{LS_{\phi}^{*}(\boldsymbol{\theta})+\epsilon}.
\end{equation}

These local metrics provide insight into the relationship between $\boldsymbol{\theta}$ and its neighbors. Solutions with relatively small $r^{*}(\boldsymbol{\theta})$ values are analogous to local optima in classical fitness landscapes - due to their large (in relative terms) expected local sensitivity they are likely to be selected in NS-like algorithms, despite their low expected evolvability. Hence, we suspect that they could represent points of deception and that their presence is expected to ramp up the difficulty of the respective behavior landscape. We refer to such solutions as \textbf{"dissimila"}, in analogy to "optima". On the contrary, solutions with relatively large $r(\boldsymbol{\theta})^{*}$ values represent highly evolvable points in the behavior landscape that might be difficult to find because of their relatively low expected local sensitivity, i.e. their relative similarity to their neighbors.

\subsection{Walks}

A \textbf{walk} \cite{stadler2002fitness} $\mathcal{W}$ on a behavior landscape is a sequence (time series) of behavior vectors $(\mathbf{b_0}, \ldots, \mathbf{b_i}, \ldots, \mathbf{b_T})$, associated with a sequence of genotypes, $(\boldsymbol{\theta_0}, \ldots, \boldsymbol{\theta_i}, \ldots, \boldsymbol{\theta_T})$, via the relation $\boldsymbol{b_{i}} = \phi(\boldsymbol{\theta_{i}})$. The underlying sequence of genotypes is generated by some process $\mathcal{A}$ (e.g. an evolutionary algorithm), i.e., $\mathcal{A} (\boldsymbol{\theta_i})=\boldsymbol{\theta_{i+1}}$. This ensures that actual segments of the search space traversed by the respective (novelty search) algorithm are considered \cite{rodrigues2020study}. Solution $\boldsymbol{\theta_{i+1}}$ is called the child of $\boldsymbol{\theta_{i}}$, and conversely, $\boldsymbol{\theta_{i}}$ is referred to as the parent of $\boldsymbol{\theta_{i+1}}$. Unless stated otherwise, real-valued (non-categorical) BVs are assumed. Also, all walks are assumed to be of finite length $T+1$. We highlight that walks are particularly well-suited to evolvability analysis, as they involve a single parent generating a large number of offspring (neighbors) during each step.

\subsubsection{Highly selective walk}

In a highly selective walk, the best (according to some criterion) solution from $\mathcal{S}(\boldsymbol{\theta}_i)$ is chosen, i.e.:
\begin{equation} \label{eq:2}
    \mathcal{A} (\boldsymbol{\theta_i}) = \boldsymbol{\theta_{i+1}} = \underset{\boldsymbol{\theta'} \in \mathcal{S}(\boldsymbol{\theta}_i)}{\mathrm{argmax}} \mathcal{G} (\phi(\boldsymbol{\theta'})),
\end{equation}
where $\mathcal{G}$ is a (novelty-based) fitness metric and $\mathcal{S}(\boldsymbol{\theta}_i) \subseteq \mathcal{N}(\boldsymbol{\theta_i})$ is defined as earlier. This type of walk represents situations in which there is a large amount of evolutionary pressure. If niches are considered, an alternative definition is possible, given by:
\begin{equation} \label{eq:2}
    \mathcal{A} (\boldsymbol{\theta_i}) = \boldsymbol{\theta_{i+1}} = \underset{\substack{\boldsymbol{\theta}' \in \mathcal{S}(\boldsymbol{\theta}_i) \\ N(\phi(\boldsymbol{\theta}')) \neq N(\phi(\boldsymbol{\theta}_{i}))}}{\mathrm{argmax}} \mathcal{G} (\phi(\boldsymbol{\theta'})),
\end{equation}
where $N(\phi(\boldsymbol{\theta}))$ denotes the niche to which $\phi(\boldsymbol{\theta})$ belongs. Hence, among solutions that fall into a different niche than $\boldsymbol{\theta_i}$, the best one is accepted. The corresponding walk through niches is given by: $(N(\phi(\boldsymbol{\theta_{0}})),  \ldots, N(\phi(\boldsymbol{\theta_{i}})), \ldots, N(\phi(\boldsymbol{\theta_{T}})))$.

\subsubsection{Adaptive walk}
In an adaptive walk, any solution "better" than $\boldsymbol{\theta_{i}}$ is accepted. It is an intermediate type of walk between a highly selective and a random walk. Consider first the subset of children of $\boldsymbol{\theta}_i$, given by $C_{a} = \{ \boldsymbol{\theta}' \in \mathcal{S}(\boldsymbol{\theta}_i) \mid \mathcal{G} (\phi(\boldsymbol{\theta}') \geq a \}$ for some $a$ which is a proxy for evolutionary pressure. Depending on the model, $a$ can be constant or change dynamically in time. Now simply:
\begin{equation}
    \boldsymbol{\theta_{i+1}} \sim \text{Unif}(C_a),
\end{equation}
where $\text{Unif}$ denotes the uniform distribution.
Hence, unlike the case with highly selective walks, a random solution that surpasses a certain threshold is selected instead. If no such solution exists, set $\boldsymbol{\theta}_{i+1}=\boldsymbol{\theta}_{i}$ and sample from $\mathcal{S}(\boldsymbol{\theta}_{i+1})$ again. Let us also consider adaptive walks when using BVs expressed as one-hot vectors. In that case, consider the subset $C_{a} = \{ \boldsymbol{\theta}' \in \mathcal{S}(\boldsymbol{\theta}_i) \mid N(\phi(\boldsymbol{\theta}')) \neq N(\phi(\boldsymbol{\theta}_i)) \}$ and sample uniformly from it to obtain $\boldsymbol{\theta_{i+1}}$. Similarly as before, if $C_a$ is an empty set, let $\boldsymbol{\theta_{i+1}}=\boldsymbol{\theta_{i}}$.
We emphasize that a highly selective walk can also be considered adaptive, with $a$ selected in such a way that $\mathbf{card}(C_a)=1$. Finally, we raise attention to the connections between biological evolution by natural selection and the concept of adaptive walks \cite{gillespie1984molecular, kauffman1987towards}.

\subsubsection{Random walk} In a random walk, any neighbor is accepted:
\begin{equation}
    \boldsymbol{\theta_{i+1}} \sim \text{Unif}(\mathcal{S}(\boldsymbol{\theta}_i)).
\end{equation}
It describes the conditions under which there is no evolutionary pressure altogether.

\subsubsection{Multi-state walk}

In a multi-state walk, at each step $i$, a set of genotypes is selected, resulting in a matrix $\mathcal{Z}_i$ filled with BVs. 
For the sake of simplicity, in what follows the focus is solely on single-state walks. By doing this we abstract the key aspects of divergent search, stripping it to its barest minimum and focusing on the evolvability of individuals.

\subsection{Diversity metrics}

There are various options available when selecting the diversity metric $\mathcal{G}$. Here we list and propose several:
\subsubsection{K-nearest neighbors distance to other individuals in the population and the archive}
This case corresponds to the vanilla novelty search algorithm \cite{lehman2011novelty}:
\begin{equation} \label{eq:5}
    \mathcal{G} (\phi(\boldsymbol{\theta'})) =  \sum_{1 \leq k \leq K} |\mu_k(\boldsymbol{\theta}')-\phi(\boldsymbol{\theta}')|
\end{equation}
where $\mu_k(\boldsymbol{\theta'})$ is the phenotype of the $k$-th nearest (in the BS) individual to $\boldsymbol{\theta}'$, with $\mu_k(\boldsymbol{\theta'}) =  \phi(\boldsymbol{\theta}'')$, $\boldsymbol{\theta}'' \in \mathcal{S}(\boldsymbol{\theta}_i) \cup \mathcal{X}$. $\mathcal{X}$ is the archive of previous novel (or randomly selected) individuals. The Euclidean distance is typically used.
\subsubsection{Distance to ancestors}
If the distance to the parent is used as a metric, we have:
\begin{equation} \label{eq:3}
    \mathcal{G} (\phi(\boldsymbol{\theta'})) =  |\phi(\boldsymbol{\theta}_i)-\phi(\boldsymbol{\theta}')|.
\end{equation}
Observe the similarity to Eq. \ref{eq:1}; if $\mathcal{S}(\boldsymbol{\theta}_i) = \mathcal{N}(\boldsymbol{\theta_i})$, the distance between consecutive BVs in the respective walk is maximum and equal to the local sensitivity at $\boldsymbol{\theta}'$. Alternatively, it is possible to consider not only $\boldsymbol{\theta_{i}}$ but also the entire chain of ancestors, which then acts as a simplified archive. More formally:
\begin{equation} \label{eq:44}
       \mathcal{G} (\phi(\boldsymbol{\theta'})) = \sum_{0 \leq j \leq i} |\phi(\boldsymbol{\theta}_j)-\phi(\boldsymbol{\theta}')|.
\end{equation}
The Markovian property no longer holds with Eq. \ref{eq:44}.

\subsubsection{Gaussian kernel density estimation-based metric.} Finally, it is possible to set:
\begin{equation} \label{eq:45}
       \mathcal{G} (\phi(\boldsymbol{\theta'})) =  \frac{1}{\mathbf{card}(\mathcal{Y})} \sum_{\boldsymbol{\theta}'' \in \mathcal{Y}} -K_{\boldsymbol{H}} \left( \phi(\boldsymbol{\theta'})-\phi(\boldsymbol{\theta''})\right) \gamma(\boldsymbol{\theta''}),
\end{equation}
where $\mathcal{Y} = (\mathcal{S}(\boldsymbol{\theta}_i) \cup \mathcal{X}) \setminus \{ \boldsymbol{\theta}' \} $, $K_{\mathbf {H} }$ is the Gaussian kernel, $\boldsymbol{H}$ is the bandwidth\footnote{Also known as the smoothing matrix.} matrix and the rest of the notation is the same as before. The Gaussian kernel is given by:
\begin{equation}
    K_{\mathbf {H}}(\mathbf {x} )={(2\pi )^{-d/2}}\mathbf {|H|} ^{-1/2}e^{-{\frac {1}{2}}\mathbf {x^{T}} \mathbf {H^{-1}} \mathbf {x}},
\end{equation}
with $d$ denoting the number of dimensions. The idea is to select solutions that are located remotely from the areas already explored. However, since visiting previously seen regions (i.e., backtracking) can even be beneficial \cite{salehi2022geodesics}, we introduce a discount function $\gamma$ that ensures that recently explored regions are avoided more strongly than those visited long ago, as well as use an archive with a finite size. In the context of biological evolution, this approach could resemble a group of individuals searching for food while avoiding areas depleted of resources by previous inhabitants, in an environment that slowly replenishes resources over time. Note that by using such a metric, the archive is implicitly contained in the shape of the kernel density estimate, which plays the role of a memory container. Also, observe that the choice of the matrix $\boldsymbol{H}$ is a subtle one, as it determines the amount of smoothing applied to the density estimate, with small (large) bandwidths resulting in more peaked (smoother) estimates. 

\section{Experimental setup}

\begin{figure}
  \centering
  \includegraphics[width=0.35\linewidth]{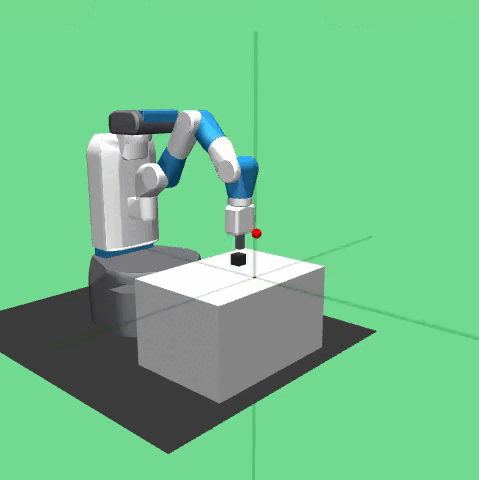}
  \caption{Visualization of the Fetch "Pick and Place" environment rendered in observable mode.  A robotic arm is shown moving around the table, about to grab the block.}
  \label{fig:robot}
\end{figure}

Experiments on an evolutionary robotics task are conducted to examine the effect of evolutionary pressure and the use of different diversity metrics on evolvability. Although the used environment in its original form includes an explicit goal, e.g. moving a block to a target location, our study focuses solely on the diversity of learned behaviors and disregards any other goals. The experiments are implemented using \verb|Gymnasium| \cite{1606.01540}, an open-source Python library for reinforcement learning, and its associated collection of environments, \verb|Gymnasium Robotics|. Parallel processing is achieved with \verb|Ray|, a toolkit for scaling Python applications to clusters. The environment used for the experiments is illustrated in Fig. \ref{fig:robot}.

\subsection{Task description}


The experiments are performed on a toy episodic reinforcement learning task provided within the Fetch "Pick and Place" environment \cite{1802.09464}. The environment is based on the Fetch Mobile Manipulator, a seven-degree-of-freedom robotic arm equipped with a two-fingered parallel gripper as its end-effector, which can be either closed or open. The robot is controlled by moving the gripper in a Cartesian space. In the original task, the goal is to use the manipulator to pick up and move a block from an initial position to a target position. The block is initially located on a table, while the target position can either be on the table or in mid-air. In the task formulation with rewards, a positive reward is given upon completing the goal (if using a sparse reward formulation) or moving the block closer (if using a dense reward formulation) to the target position. The manipulator is controlled at a frequency of $f = \SI{25}{\hertz}$ and each simulation time-step has a duration of $dt = \SI{0.002}{\second}$. The observation space consists of $25$ variables, including the current positions, displacements, rotations, and velocities of various parts of the end-effector and the block. Kinematic information is provided through Mujoco \cite{todorov2012mujoco} bodies attached to both the block and the end-effector. The action space consists of four variables: Cartesian displacements $dx$, $dy$, and $dz$ of the end-effector, and one variable controlling the closing and opening of the gripper. The episode is truncated when its duration reaches the maximum number of steps, which is set to $50$ by default. The determinism of the environment is ensured by fixing the initial conditions, more specifically, the initial and the target position of the block.

\subsection{Behavior space and descriptors}

In the considered environment, the BS is two-dimensional and the BVs are given by $\mathbf{b} = (x_T, y_T)$ where $x_T$ ($y_T$) denotes the final position of the block in the $x$ ($y$) axis direction.

\subsection{Neural network controllers}

A fully-connected feed-forward neural network architecture was used for the experiments. The first layer consists of $25$ input neurons that encode the state (observation) space features. It is followed by two hidden layers, each with $32$ neurons.  The output layer contains only $4$ neurons (the same as the cardinality of the action space) which are responsible for controlling the robotic manipulator. The rectified linear unit (\verb|ReLU|) activation function was used in all places, with the exception of the output layer, where the hyperbolic tangent (\verb|tanh|) activation function was employed to ensure that the actions produced by the network are within the $[-1,1]$ range, as required by the used reinforcement learning environment. The choice of the neural network architecture was guided partly by prior works that showed the effectiveness of relatively shallow designs (with only 2 hidden layers) on various reinforcement learning tasks with low-dimensional state spaces \cite{kurek2016heterogeneous}, and partly by our initial experimentation. The resulting policies are deterministic in the sense that states are mapped to specific actions as opposed to probability distributions over the action space. The Xavier normal\footnote{Weights are sampled from a normal distribution with mean $0$ and variance inversely proportional to the sum of the number of inputs and outputs to the layer.} initialization was used for the weights, while the biases were initialized to a constant value of $0$. The topology of the neural network remains constant, and only the weights are updated during the search.

\subsection{Variation operator}

The choice of mutation operator is pivotal as it defines the set of neighbors and the probability distribution over it, thereby affecting the search through the behavior landscape. We use the standard Cauchy distribution (also known as the Student's t-distribution with a single degree of freedom) to perturb the weights of the employed neural network. For simplicity, the crossover is not utilized in our experiments. This choice was partly inspired by recent work in which an evolutionary algorithm based on the use of Cauchy deviates (simple Cauchy mutations) was shown to have favorable convergence properties \cite{bajer2016population}, although this was done outside of the context of neuroevolution. It should be noted that, due to the distribution's leptokurtosis, the use of Cauchy-based mutations tends to frequently cause small changes to the respective genotype, and occasionally large ones. 

\subsection{Hyperparameters}

The size of the archive is limited, and new solutions are randomly added to it with a probability of $P = 10 \% $. The values for the remaining hyperparameters can be found in Table \ref{tab:freq} and in the captions of the respective figures. The selection of certain hyperparameters, such as the number of walks (run) or the walk length, was made in such a way as to ensure that the simulations were manageable within the budget of our available computational resources. Additionally, the parameter $a$ (required for generating adaptive walks) is set dynamically such that in each generation, the top x$\%$ of individuals pass the threshold.

\begin{table}
  \caption{Hyperparameter values}
  \label{tab:freq}
  \begin{center}
  \begin{tabular}{cl}
    \toprule
    Hyperparameter&Value\\
    \midrule
    Num. of offspring per generation & 30\\
    Smoothing parameter & 0.5\\
    Archive size limit & 1200\\
    Num. of walks (runs) per configuration & 50\\
    Num. of nearest neighbors $k$ & 15\\
  \bottomrule
\end{tabular}
\end{center}
\end{table}

\section{Results}

\subsection{Evolutionary pressure}

\begin{figure}
  \centering
  \includegraphics[width=0.6\linewidth]{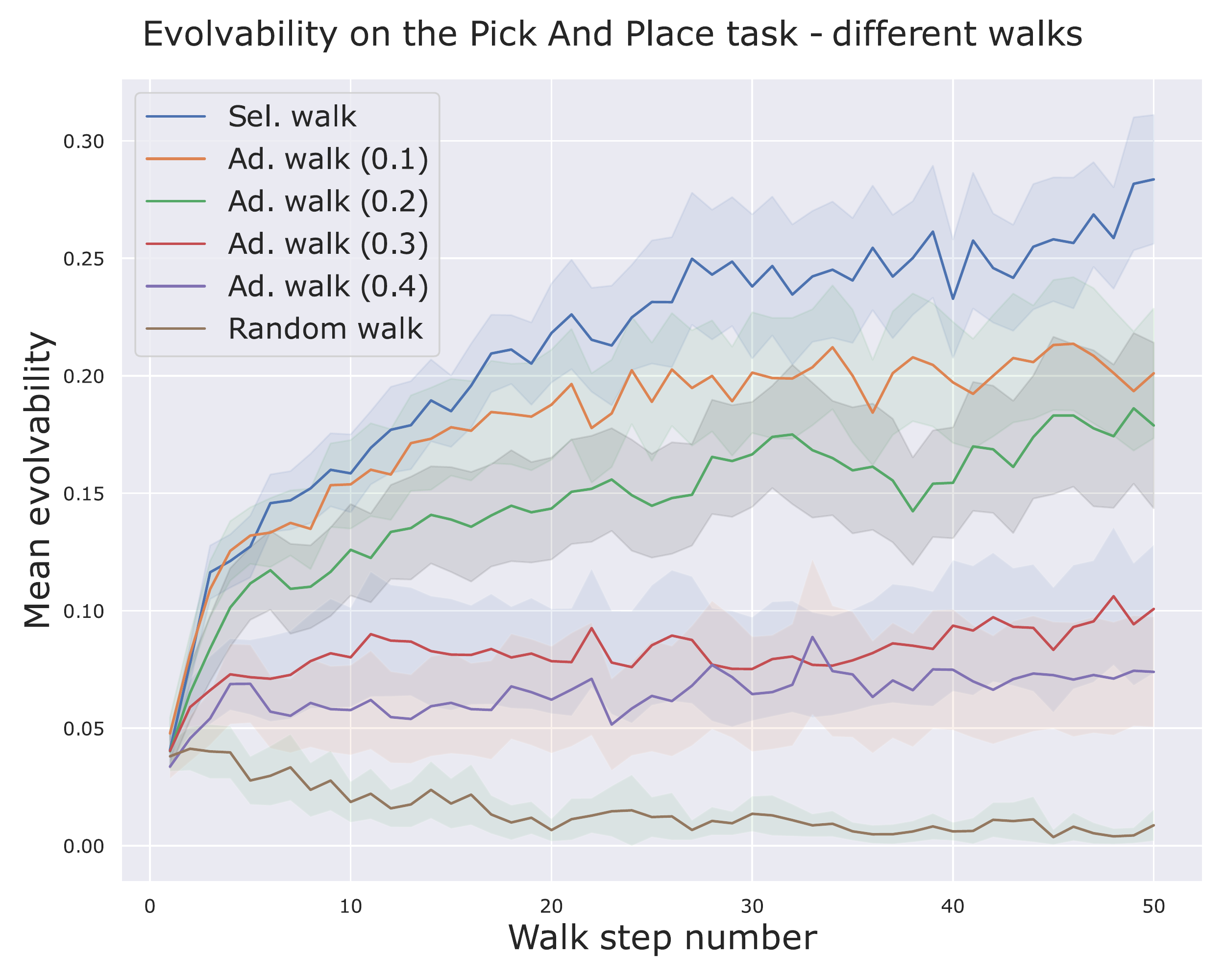}
  \caption{Comparison of mean evolvability for selective (sel.), adaptive (ad.), and random walks on the behavior landscape of the Fetch Pick and Place environment. The figure displays the means and $95 \%$ confidence intervals for a fixed number of task runs.}
  \label{fig:noises}
\end{figure}

We start by analyzing the effect of evolutionary pressure on evolvability. Fig. \ref{fig:noises} shows the mean evolvability under a fixed number of runs as a function of the walk step number, for walks with different levels of evolutionary pressure. The number of walk steps is set to $50$. Since highly selective walks instigate evolvability, adaptive walks employing novelty as a criterion are also expected to do this, albeit to a smaller degree. The parameter (given in parentheses) specifies the fraction of the best children that comprise the set from which the next parent is randomly picked. Thus, higher values indicate lower degrees of evolutionary pressure. Extremes are represented by the highly selective walk, in which the best single child is always selected, and the random walk (in essence a random search in the genotype space) in which any child is chosen. Observe that, after starting from approximately the same mean evolvability value, the trajectories spread out in a manner that suggests a positive correlation between evolvability and the level of evolutionary pressure. We use evolvabilities at the end of the walk (step $50$) and the corresponding levels of evolutionary pressure to determine the Spearman rank-order correlation coefficient $r$, which is a measure of the monotonicity of the relationship between two variables. The obtained correlation coefficient equals $r = 0.7106$ ($p < 10^{-46})$, demonstrating a strong positive correlation. Also note that, while the random walk trajectory leads to a clear degradation of evolvability over time, even modest amounts of evolutionary pressure are sufficient to preserve (or even enhance) it for all of the considered trajectories. In summary, the obtained results are in accordance with recent findings on novelty search promoting evolvability \cite{doncieux2020novelty} and also offer additional perspective on the relationship between evolvability and evolutionary pressure. We also leave for further work the following question: \textit{Is it possible to devise neural network initialization schemes that lead to high levels of evolvability at the very beginning, introducing one-shot evolvability and facilitating the ensuing search process?} This is hoped to lead to even faster evolvability gains.

\subsection{Different metrics}

\begin{figure}
  \centering
  \includegraphics[width=0.6\linewidth]{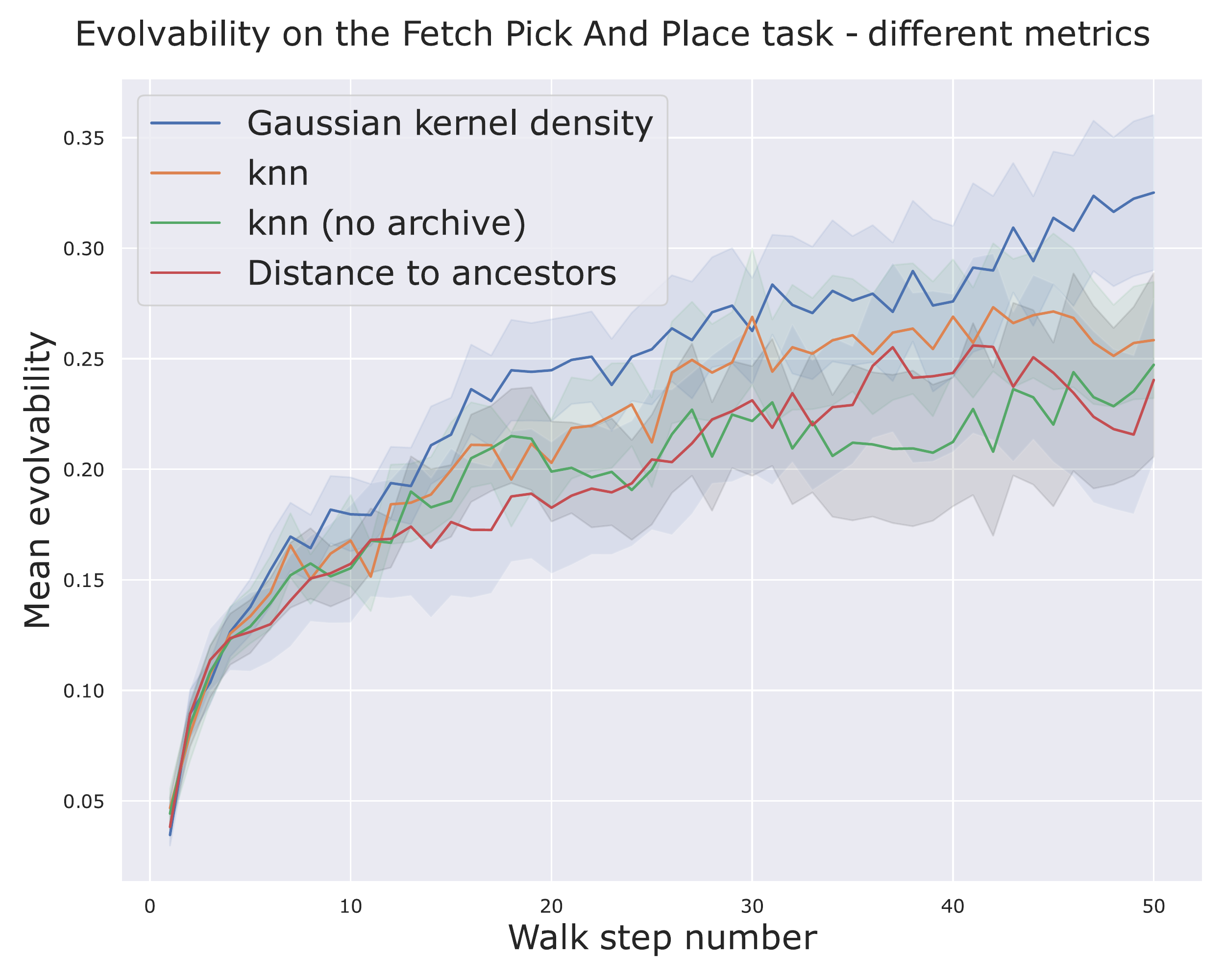}
  \caption{Comparison of different metrics on the behavior landscape of the Fetch Pick and Place environment. Means and confidence intervals $(95 \%)$ are shown for a fixed number of runs on the task. The $\gamma$ discount factor associated with the Gaussian kernel density-based metric is simply set to $1$.}
  \label{fig:noises3}
\end{figure}

In this Subsection, we analyze the effect of different diversity metrics on evolvability. Fig. \ref{fig:noises3} shows the results on the Fetch "Pick and Place" environment, with the number of walk steps set to $50$. We emphasize that the distance to ancestors metric takes into account the whole chain of ancestors, up until the very beginning of the walk. As seen in the figure, the metric based on Gaussian kernel density estimation seems to dominate over the alternatives, while the k-nearest neighbors without the archive variant yields the lowest evolvability values. To corroborate this observation statistically, consider first the sets of evolvability values at walk step $100$ associated with each of the evaluated metrics. To this end, we turn to the non-parametric Kruskal-Wallis H-test, which does not require the assumption of normality, while its other assumptions (independence of samples, random sampling, ordinal or continuous dependent variable) are met. It is then used to test the null hypothesis that the population medians of all of the considered sets are equal. The obtained p-value equals $p=0.01943$, rejecting the null hypothesis at $\alpha=0.05$ significance level. The Conover's test is used as a post-hoc test to make pairwise comparisons, with the choice of the Holm method for adjusting p-values. The p-values comparing the Gaussian kernel density estimates to (a) the KNN method, (b) KNN without an archive, and (c) the distance to ancestors-based metric are, in order: $0.053031$, $0.052218$, and $0.040418$. All the other p-values are equal to $1$. Hence, given the $\alpha=0.05$ significance level, we cannot conclude that there is a significant difference between the mean ranks of these groups, except for the Gaussian kernel density estimates/distance to ancestors pair. Nevertheless, the results do indicate that the use of the Gaussian kernel density estimate-based metric provides a competitive, if not superior, alternative that could be added to the arsenal of existing metrics. An important difference between the Gaussian kernel density estimate-based metric and the one based on KNN with Euclidean distance should be highlighted. Consider, for example, points $A$ and $B$ in a BS and the line connecting them. KNN with Euclidean distance (assuming $k=2$) would be indifferent between the points on the line, which represents a contour line in the corresponding contour graph. In contrast, a Gaussian kernel density estimate-based metric would favor the mid-point.

While the results are so far limited to a single reinforcement learning environment, further work should test generalization across multiple environments. Finally, keep in mind that highly selective walks represent a very simplified version of novelty search with only a single parent in each generation. Consequently, it is possible and even likely that the presented results underestimate the evolvability levels that would appear in a fully-fledged novelty search algorithm. 

\section{More on evolvability}
\subsection{Towards better evolvability definitions}

It has been observed, already in the original MAP-Elites paper \cite{mouret2015illuminating}, that most solutions are positioned in the BS relatively close to their parents \cite{mouret2015illuminating}. Donciex \emph{et al.} \cite{doncieux2020novelty} measure evolvability by estimating reachability (with a simple ratio) and uniformity separately (by using the Jensen-Shannon distance). The authors emphasize that separating the two aspects facilitates interpretation and prevents informational overlap, as the same Jensen-Shannon Distance (JSD) can result from sets of points with varying coverage. While valuable, we argue that this separation alone does not deal with the central problem because two sets of points with identical coverage and JSD can greatly vary in terms of their actual evolvability. More specifically, the ensuing definition is highly myopic in the sense that it does not take into account the evolvability of the offspring themselves. To illustrate this, consider from example solutions $A$ and $B$, each with two children, belonging to niches $1$ and $2$, and $3$ and $4$, respectively. $A$ and $B$ clearly have the same coverage and JSD. However, if solutions from niches $3$ and $4$ are more likely to generate offspring belonging to other niches than solutions from niches $1$ and $2$, it could be argued that $B$ should be considered more evolvable. Furthermore, the evolvability of the visited niches should also be accounted for, some of which might serve as "wormholes", generating offspring in remote or hardly reachable regions of the BS. Therefore, evolvability metrics should consider not only the coverage and uniformity but also the very structure of the behavior landscape, i.e. the mutual reachability of different niches (regions) in it and the very structure of the genotype-phenotype mapping. With this goal in mind, we suggest the following criterion, which presents a generalization, applicable when the BS is split into niches: \\
\textit{Evolvability in discretized BSs:} A solution $\boldsymbol{\theta}'$ is said to be more $l$-evolvable than solution $\boldsymbol{\theta}''$ if and only if: 
\begin{equation}
\mathbb{E}(\mathbf{card}(\mathcal{D}_{\boldsymbol{\theta}'}^{l,\ U})) > \mathbb{E}(\mathbf{card}(\mathcal{D}_{\boldsymbol{\theta}''}^{l,\ U})),
\label{def:exp}
\end{equation}
where $\mathbf{card}(\mathcal{D}_{\boldsymbol{\theta}}^{l,\ U})$ denotes the cardinality of the set of all niches visited at least once by descendants of ${\boldsymbol{\theta}}$ up to their $l$-th generation, assuming a total of $U$ descendants. The parameter $l$ captures the level of long-sightedness, with $l=1$ corresponding to the simple coverage of the children, as introduced in \cite{doncieux2020novelty}. In non-discretized BSs, the number of visited niches could be replaced, for example, by the hypervolume of the convex hull spanned by the descendants. In what follows, we lay out a procedure for estimating the evolvability of solution $\boldsymbol{\theta}$ under such a definition:

\begin{enumerate}
\item Use a grid to split the BS into niches
\item Generate the offspring of $\boldsymbol{\theta}$ and calculate their discrete distribution $D$ over the set of all niches
\item Use MAP-Elites \cite{mouret2015illuminating} or other NS-like algorithm to estimate the probability matrix $T=(t_{i, j})$, where $t_{i, j}$ is the probability of solution in niche $i$ generating a child in niche $j$
\item Construct a Markov chain with $T$ as its transition matrix, the set of niches as its state space, and $D$ as its initial distribution
\item Perform $U$ random walks of length $l+1$ on the constructed Markov chain and obtain the discrete distribution $D''$ of the visited niches over the set of all niches, cumulatively for all $U$ walks 
\item Apply an evolvability metric to $D''$, i.e. coverage (in accordance with Expression \ref{def:exp}) 
\item Repeat steps (5) and (6) multiple times, and use the mean to approximate the evolvability of $\boldsymbol{\theta}$
\end{enumerate}

Additionally, it is possible to define the evolvability of niche $i$ by using the appropriate initial distribution (i.e. a one-hot vector) as $D$. We remark on several aspects. Firstly, selecting the correct number of niches (i.e. the grid size) in the first step is a delicate matter, as having too few niches could result in the matrix $T$ with mostly close-to-zero values in all its non-diagonal entries, while having too many can lead to a large number of unvisited niches. Secondly, note the similarity with the curiosity score \cite{cully2017quality, gravina2018quality} used in quality-diversity approaches. The shortcoming of the method (in the presented form) lies in the fact that it can only be used \emph{a posteriori}, i.e. after the entire BS has already been explored. Despite this limitation, it provides a useful tool for better exploring the structure of the behavior landscape, primarily through the estimation of the transition matrix $T$. Also, by evaluating $T$ in an online fashion, it could also be integrated with other methods that search directly for evolvability \cite{mengistu2016evolvability}.

\section{Conclusion and further work}

This paper explores evolvability in neuroevolutionary divergent search from the perspective of a behavior landscape analysis. In particular, the results confirm the positive correlation between evolvability and the amount of evolutionary pressure present in the environment. Moreover, they indicate that the use of Gaussian kernel density estimate-based metrics should be added to the arsenal of existing diversity metrics in the context of evolvability promotion. Finally, a new definition of evolvability, that considers the offsprings' evolvability as well and takes into account the very structure of the behavior landscape, is proposed, along with a method for its estimation.

As part of further work, generalization to a wider range of environments should be studied, along with the use of alternative operators and behavior space descriptors. The relationship between the newly proposed definition of evolvability and the curiosity score should also be experimentally studied. Another possible path forward is to investigate the links between the evolvability of neural network controllers and their sensitivity to initial conditions in stochastic (noisy) reinforcement learning environments, i.e., a form of Lyapunov stability \cite{perkins2002lyapunov}. More specifically, the phenotypic sensitivity to environmental noise (with a fixed genotype) could be used as a behavior (meta)-descriptor. Further ideas rely on the direct evolvability search \cite{mengistu2016evolvability} and lead it into new directions. One idea would be to take it a step further and consider direct meta-evolvability search: instead of optimizing for evolvability, optimize for the potential for evolvability. Another possible avenue is to encode the parameters of the mutation operator's distribution into the genotype, thereby also performing a search through the space of heuristics. More generally, the use of hyperheuristics such as genetic programming might help uncover heuristics with the capacity to further enhance evolvability. Further studies could also test the use of Gaussian kernel density estimate-based distance metrics in the context of population-level evolvability \cite{lehman2016critical}. Finally, the effects of optimization with conflicting objectives \cite{smith2016exploring} on evolvability could also be considered.

\bibliographystyle{unsrt}
\bibliography{references}  

\end{document}